\documentclass{bmvc2k}

\usepackage{graphicx}
\usepackage{amsmath}
\usepackage{amssymb}
\usepackage{tabularx}
\usepackage{pdfpages}
\usepackage{booktabs}
\usepackage{subcaption}
\usepackage[option]{colortbl}
\usepackage{wrapfig}
\usepackage{lipsum}
\usepackage{comment}
\usepackage{hyperref}
\newcommand{\ryu}[1]{{\color{black}{#1}}}

\newcommand{\cRyo}[1]{{\color{black}{#1}}}

\newcommand\blfootnote[1]{%
  \begingroup
  \renewcommand\thefootnote{}\footnote{#1}%
  \addtocounter{footnote}{-1}%
  \endgroup
}

\title{Primitive Geometry Segment Pre-training \\ for 3D Medical Image Segmentation}

\addauthor{Ryu Tadokoro$^\ast$}{tadokororyuryu@gmail.com}{1,2}
\addauthor{Ryosuke Yamada$^\ast$}{ryosuke.yamada@aist.go.jp}{1,3}
\addauthor{Kodai Nakashima}{nakashima.kodai@aist.go.jp}{1,3}
\addauthor{Ryo Nakamura}{ryo.nakamura@aist.go.jp}{1,4}
\addauthor{Hirokatsu Kataoka}{hirokatsu.kataoka@aist.go.jp}{1}

\addinstitution{
 National Institute of Advanced Industrial Science and Technology, Japan
}
\addinstitution{
 Tohoku University, Japan
}
\addinstitution{
 University of Tsukuba, Japan
}
\addinstitution{
 Fukuoka University, Japan
}

\runninghead{Tadokoro, Yamada, Nakashima, Nakamura, Kataoka}{PrimGeoSeg}

\begin{document}

\maketitle
\blfootnote{$^\ast$ These authors contributed equally to this work.}

\begin{abstract}
The construction of 3D medical image datasets presents several issues, including requiring significant financial costs in data collection and specialized expertise for annotation, as well as strict privacy concerns for patient confidentiality compared to natural image datasets. Therefore, it has become a pressing issue in 3D medical image segmentation to enable data-efficient learning with limited 3D medical data and supervision. A promising approach is pre-training, but improving its performance in 3D medical image segmentation is difficult due to the small size of existing 3D medical image datasets. 
We thus present the Primitive Geometry Segment Pre-training (PrimGeoSeg) method to enable the learning of 3D semantic features by pre-training segmentation tasks using only primitive geometric objects for 3D medical image segmentation. PrimGeoSeg performs more accurate and efficient 3D medical image segmentation without manual data collection and annotation. 
\ryu{Further, experimental results show that PrimGeoSeg on SwinUNETR improves performance over learning from scratch on BTCV, MSD (Task06), and BraTS datasets by 3.7\%, 4.4\%, and 0.3\%, respectively.}
Remarkably, the performance was \ryu{equal to or} better than state-of-the-art self-supervised learning despite the equal number of pre-training data.
From experimental results, we conclude that effective pre-training can be achieved by looking at primitive geometric objects only. 
Code and dataset are available at \href{https://github.com/SUPER-TADORY/PrimGeoSeg}{https://github.com/SUPER-TADORY/PrimGeoSeg}.
\end{abstract}

\section{Introduction}
\label{sec:intro}

3D medical image analysis using deep learning is expected to enhance diagnostics and improve patient outcomes through \cRyo{the} more accurate detection and visualization of \cRyo{geometric} structures inside the human body. For example, 3D medical image segmentation estimates the location and category of human organs from computed tomography (CT) and magnetic resonance imaging (MRI) images. 
More accurate segmentation of 3D medical images requires a large amount of training data and rich semantic annotation. 
However, training data collection is difficult due to the high imaging costs and stringent privacy protections. In addition, the annotation process requires expert knowledge of medical science. 

In order to solve the above problems, there have been many studies in terms of pre-training methods toward more data-efficient learning under limited training data conditions. In particular, self-supervised learning (SSL) has emerged as a promising approach for pre-training in 3D medical image segmentation~\cite{gibson2018niftynet, chen2019med3d,  zhou2021models, zhu2020rubik, taleb20203d,zhou2021preservational,haghighi2021transferable, ye2022desd, nguyen2022joint, you2022momentum, tang2022self, chen2023masked, jiang2022selfsupervised, xie2022unimiss}, as it learns 3D structural features and reduces manual annotation costs by designing and learning a pre-text task on unsupervised data. Chen \textit{et al.}~\cite{chen2023masked} achieved state-of-the-art performance \ryu{on the Multi-Atlas Labeling Beyond the Cranial Vault (BTCV)} dataset~\cite{BTCV} and Medical Segmentation Decathlon (MSD)~\cite{MSD} dataset by merging existing 3D medical image datasets and pre-training three pseudo tasks. Nevertheless, pre-training methods for 3D medical image segmentation have lagged compared with other 3D object recognition tasks because of the small scale of pre-training datasets. Therefore, an alternative pre-training approach is needed to address dataset construction issues in 3D medical image segmentation.

Formula-driven supervised learning (FDSL)~\cite{KataokaACCV2020, Kataoka_2021_ICCV} has been proposed as a synthetic pre-training method without real data and human annotations, which automatically generates synthetic data and supervised labels based on a specific principle rule of the real world. Therefore, a significant advantage of FDSL is that the properties of the synthetic data can be designed considering fine-tuning tasks different from real data. Furthermore, as much as possible, FDSL can reduce dataset issues related to real data, such as social bias and personal information protection. 
\ryu{Recently, Nakashima \textit{et al.}~\cite{nakashima2021can} reported that Vision Transformer (ViT) tends to focus on the outlines of an object on images in pre-training.}
\ryu{ Inspired by the above insight, Kataoka \textit{et al.}~\cite{kataoka2022replacing} proposed a Radial Counter DataBase (RCDB) that has improved the complexity of outlines and pre-training performance.}
Furthermore, Yamada \textit{et al.}~\cite{yamada2022point} proposed a Point Cloud Fractal DataBase (PC-FractalDB) based on fractal geometry to improve performance by designing 3D object detection pre-training. From these insights, we hypothesize that we can design segmentation tasks using only primitive geometric objects to achieve an effective pre-training method for 3D medical segmentation.

\begin{wrapfigure}{r}{0.55\textwidth}
    \vspace{-\baselineskip}
    \centering
    \includegraphics[width=0.99\linewidth]{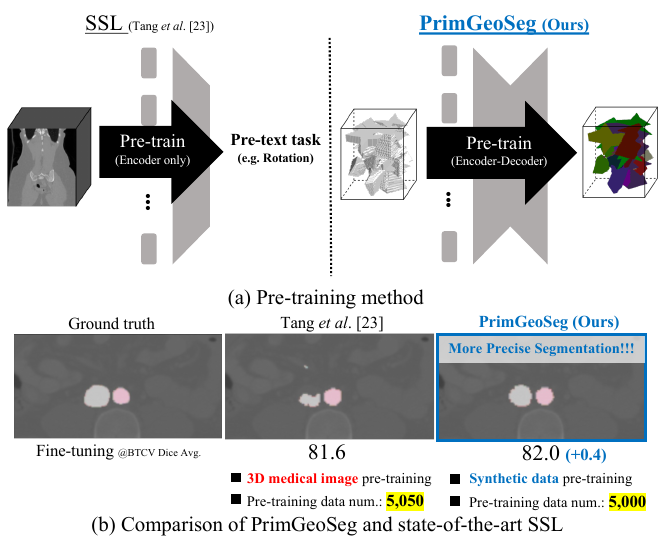}
    \vspace{-23pt}
    \caption{The overview of PrimGeoSeg.}
    \label{fig:figure1}
    \vspace{-10pt}
\end{wrapfigure}

The present study proposes a primitive geometry segmentation (PrimGeoSeg) for 3D medical image segmentation by automatically generating pre-training data and expressing semantically supervised labels as an assembly of primitive geometric objects in 3D space, as shown in Figure~\ref{fig:figure1}. 
We generate a primitive geometric object from independent laws in the $xy$-plane and $z$-axis directions. We also construct a pre-training dataset by arranging multiple primitive geometric objects in 3D space, overlapping each object.
We designed this generation process to consider two aspects of the internal structure of the human body; (i) the variability among individuals and (ii) the complexity with ambiguous boundaries between organs. The experimental results found that primitive geometric objects only are sufficient to learn the necessary 3D structural representations for achieving a superior pre-training effect for 3D medical image segmentation. 

The contributions of this work are as follows: (i) We propose PrimGeoSeg as a pre-training method that enables pre-training by segmentation without real data collection and manual annotation. (ii) We show that pre-training both UNETR and SwinUNETR with PrimGroSeg outperform state-of-the-art SSL accuracy on BTCV and MSD in 3D medical image segmentation. 
\ryu{Notably, the number of synthetic pre-training data  was almost equal (see Figure~\ref{fig:figure1}).}
In addition, PrimGeoSeg also demonstrates remarkable data efficiency, performing as well with only 30\% of the BTCV data as it does when learning from scratch with 100\% \ryu{of} training data. (iii) Our proposed method for pre-training synthetic data can reduce problems such as the privacy of 3D medical images.
\blfootnote{This paper notably extends the experiments in~\cite{Tadokoro_2023_CVPR} and provides new contributions of our proposed method.}

\section{Related Works}
\noindent{\textbf{Pre-training for 3D medical image segmentation.}} 
In 3D medical image segmentation,  SSL has been attracting attention \cRyo{for its ability to achieve} highly accurate segmentation results by pre-training unsupervised 3D medical images~\cite{gibson2018niftynet, chen2019med3d,  zhou2021models, zhu2020rubik, taleb20203d, zhou2021preservational, haghighi2021transferable, ye2022desd, you2022momentum, tang2022self, chen2023masked,chen2020improved, jiang2022selfsupervised, xie2022unimiss}.
Even in transformer-based models that achieve higher accuracy than conventional CNN-based models for 3D medical images~\cite{hatamizadeh2022unetr, hatamizadeh2022swin}, SSL \cRyo{has} shown substantial accuracy improvements. 
Chen \textit{et al.}~\cite{chen2023masked} improved performance on UNETR through pre-training \cRyo{via} masked image modeling, which masks a portion of 3D medical images. 
\cRyo{In addition}, Tang \textit{et al.}~\cite{tang2022self} achieved state-of-the-art results using the SwinUNETR~\cite{hatamizadeh2022swin} on BTCV~\cite{BTCV} and MSD~\cite{MSD} datasets by pre-training three pre-text tasks including image inpainting, 3D rotation prediction, and contrastive learning.
As shown above, SSL can improve \cRyo{the performance of 3D image medical segmentation}.
However, SSL improvements may be limited by training data \cRyo{available}, as SSL performance is often dependent on the amount of training data. \cRyo{We thus believe that the performance of the pre-training of 3D medical image segmentation will be further improved by solving the dataset construction issues.}

\noindent{\textbf{Formula-driven supervised learning (FDSL).}} 
Recently, large-scale pre-training has made tremendous \cRyo{developments} in computer vision~\cite{kolesnikov2020big, goyal2021self, beal2022billion}, \cRyo{and among its}  methods, FDSL \cRyo{can} perform large-scale pre-training without real data and manual annotation~\cite{KataokaACCV2020, Kataoka_2021_ICCV, kataoka2022replacing, nakashima2021can, takashima2023visual, KataokaWACV2022, inoue2021initialization}. Specifically, pre-training data and its label are automatically generated from mathematical formulations based on real-world principles, such as fractal geometry and Perlin noise. 
Kataoka \textit{et al.}~\cite{kataoka2022replacing}, proposed RCDB inspired that ViT pays attention to the outer contours of the fractal region \cRyo{when pre-training with the Fractal Database (FractalDB).} 
RCDB pre-trained model surpasses the ImageNet pre-trained model on ViT despite not learning natural images. More recently, Yamada \textit{et al.}~\cite{yamada2022point} proposed the PC-FractalDB for pre-training in 3D object detection using 3D point clouds. 
They \cRyo{concluded} that one factor for success in pre-training is \cRyo{initializing} not only the backbone network but also the entire model.

Based on these findings, we hypothesize that synthetic pre-training through the same segmentation task\cRyo{, similar to the} fine-tuning task, will have \cRyo{a greater} effectiveness in \cRyo{3D medical image segmentation} using the transformer-based model. Furthermore, we think that learning from synthetic pre-training data, rather than from 3D medical images, can effectively address several issues commonly associated with 3D medical data usage. These issues include societal bias, privacy concerns, and copyright infringement.


\section{PrimGeoSeg: Primitive Geometry Segment Pre-training}
\label{AVS}
In this section, we introduce PrimGeoSeg method, which is the pre-training strategy of generating primitive geometric objects and performing segment pre-training for downstream tasks in 3D medical image segmentation. 
We generate an assembled object as pre-training data for PrimGeoSeg based on the design concept of the property of 3D medical images by (i) the variability among individuals and (ii) the complexity with ambiguous boundaries between organs. 
The pre-training dataset consisting of assembled objects and supervised labels denoted by $\mathcal{D} = \{(S_{i}, m_{i})\}_{i=1}^{N}$, where $S_{i} \in \mathbb{R}^{W \times H \times D}$ is an assembled object, $m_{i} \in \mathbb{L}^{W \times H \times D}$ is a corresponding segmentation mask, and $N$ is the number of pre-training data for PrimGeoSeg. {\color{black} $\mathbb{L}$ is a set of integers denoting the segmentation label.} 

\begin{figure*}[t]
  \centering
   \includegraphics[width=1\linewidth]{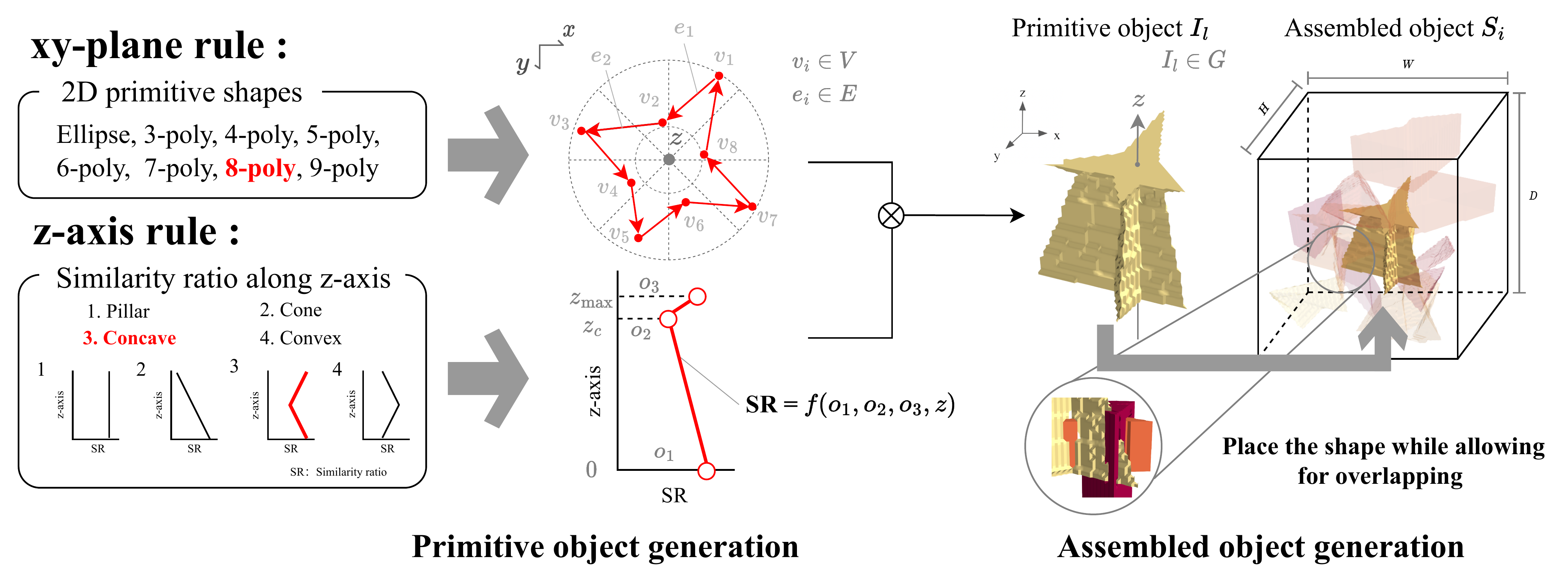}
   \caption{{\textbf{The generation process of assembled objects as pre-training data for  PrimGeoSeg.} We generate an assembled object by arranging randomly multiple primitive objects generated from the individual $xy$-plane and $z$-axis rules. 
   }}
   \label{fig:main}
   \vspace{-12pt}
\end{figure*}

As shown in Figure~\ref{fig:main}, the generation procedure of an assembled object is composed of two steps: (i) primitive object generation and (ii) arrangement of primitive objects. (i) First, we set a class of each primitive object based on $xy$-plane and $z$-axis rules. Moreover, we generate a primitive object based on randomly determined parameters regarding the number of vertices in the $xy$-plane and the similarity ratio along the $z$-axis. (ii) Second, in the arrangement of primitive objects, we generate an assembled object $S_{i}$ and its corresponding segmentation mask $m_{i}$ by arranging multiple primitive objects in 3D space. Finally, we repeat (i) -- (ii) steps $N$ times to automatically construct pre-training dataset $\mathcal{D}$ for PrimGeoSeg. 
 
\subsection{Pre-training Data Generation}
\label{sec:generation_process}

\noindent{\textbf{Primitive object generation.}} 
Each primitive object is generated by stacking $xy$-plane slices, with the similarity ratio of each slice varying along the $z$-axis.
The generation process of a primitive object is based on two rules: the $xy$-plane rule, which dictates the shape of the slices, and the $z$-axis rule, which controls the changing rate of the similarity ratio along the $z$-axis. We define a class of a primitive object considering combing the $xy$-plane and $z$-axis rules. 
We set the maximum 32 classes consisting of eight classes in the $xy$-plane rule and four classes in the $z$-axis rule (see Figure~\ref{fig:main}).
The size of the primitive object along the $z$-axis, denoted as $z_{max}$, is randomly selected from a uniform distribution, $z_{max}\sim \mathcal{U}(10,50)$. For each $t$ in the range $0 \leq t \leq z_{max}$, a slice $P_t$ is generated. 
 The similarity ratio of the slice at $z=t$ is determined by a function $f(z=t)$ according to the $z$-axis rule. 
The $z$-axis rule consists of four classes: \{`concave', `convex', `pillar', `cone'\} in this paper. 
We define $q_{z}$ by randomly selecting from the above four classes. Here, the function $f(z)$ represents the similarity ratio of the slices along with the $z$-axis direction as shown below;
\begin{equation}
f(o_1,o_2,o_3,z)=\left\{\begin{array}{l}
o_1+\left(o_2-o_1\right) \frac{z}{z_c} \quad\left(0 \leq z \leq z_c\right) \\
o_2+\left(o_3-o_2\right) \frac{z-z_c}{z_{\text {max }}-z_c}\left(z_c<z \leq z_{\text {max }}\right)
\end{array}\right.
\end{equation}
where $z_c \sim \mathcal{U}(3,z_{max}-3)$ and $o1$, $o2$, and $o3$ are defined as certain values when $q_{z}$ was selected.
For instance, in the `pillar' class, all parameters are set to $1$: $o1=o2=o3=1$. 
In the `cone' class, we use $o1=0$, $o2=\frac{z_0}{z_{max}}$, and $o3=1$. For `concave', $o1, o3 \sim \mathcal{U}(0.8,1)$, and $o2 \sim \mathcal{U}(0.2,0.5)$. For `convex', $o1, o3 \sim \mathcal{U}(0.2,0.5)$, and $o2 \sim \mathcal{U}(0.8,1)$. 
Choosing a $z$-axis rule determines the values of $o1$, $o2$, and $o3$, defining the unique function $f(z)$.

We generate the slice $P_t$ using both $f(z)$ and the $xy$-plane rule. The $xy$-plane rule is defined by a set of shape definitions: \{`ellipse', `3-poly', `4-poly', `5-poly', `6-poly', `7-poly', `8-poly', `9-poly'\}, where $w$-poly represents a $w$-sided polygon. One shape definition $q_{xy}$ is selected from the above eight rules.
To define the size of the slice $P_t$, we set parameters $R_{min}=15$ and $R_{max}\sim \mathcal{U}(30,80)$. If $q_{xy}$ is defined as a polygon, the slice $P_t$ forms a closed shape bounded by edges in the set $E(t)$:
\begin{equation}
V =\left\{\left(r_k \cos \theta_k, r_k \sin \theta_k\right) \mid 1 \leq k \leq C_{xy}\right\}
\end{equation}
\begin{equation}
E(z=t) =\left\{f(t)\left(v_k+s(v_{(k+1)\bmod C_{xy}}-v_k)\right) \mid v_k \in V, 1 \leq k \leq C_{xy}, 0 \leq s \leq 1\right\}
\end{equation}
where $r_k \sim U\left(R_{\text {min}}, R_{\text {max }}\right), \theta_k \sim U\left(\frac{2 k}{C_{xy}} \pi, \frac{2(k+1)}{C_{xy}} \pi\right)$ is polar coordinates and $C_{xy}$ is the number of vertices.
We have tolerated the alignment of three adjacent vertices in a straight line as acceptable noise. 
If $K_{xy}$ is defined as `ellipse", $P_t$ is a closed shape that satisfies the equation $ \frac{x^2}{(af(t))^2} + \frac{y^2}{(bf(t))^2} = 1 $, where $a,b\sim\mathcal{U}(R_{\text {min}},R_{\text {max}})$ is the major and minor axes of the ellipse.
By integrating the slices $P_t$ from $z=0$ to $z_{max}$ along the $z$-axis, we create a single primitive object $I_l$.
By repeating $M$ times, we generate a set of primitive objects $G = \{I_l\}_{l=1}^M$. 
Here, $M$ is the number of primitive objects \ryu{to be positioned} in assembled object $S_i$.
 
\begin{wrapfigure}{r}{0.4\textwidth}
    \vspace{-10pt}
    \includegraphics[width=1.0\linewidth]{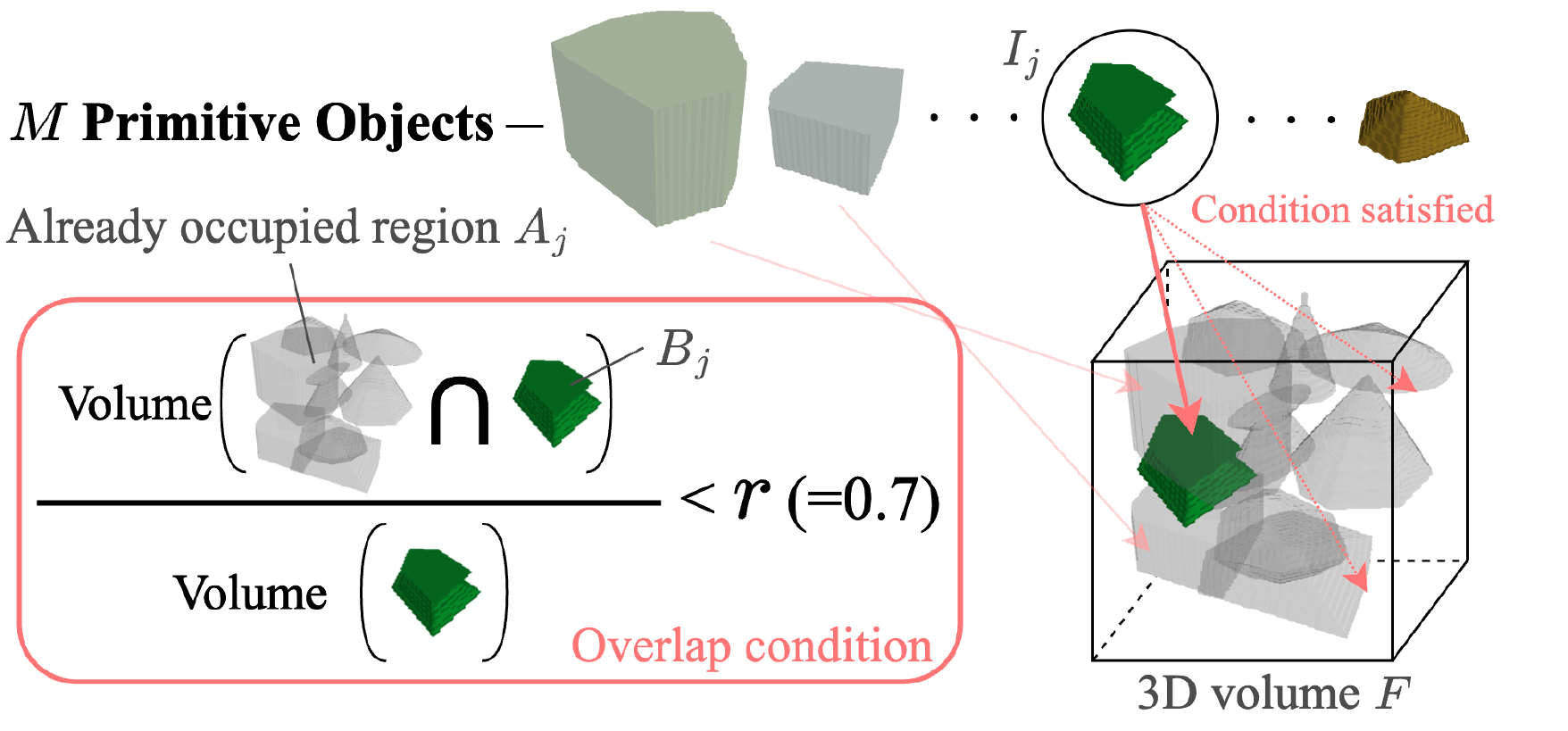}
    \vspace{-20pt}
    \caption{The details of the arrangement of primitive objects.}
    \vspace{-15pt}
    \label{fig:arrangement}
\end{wrapfigure}

\noindent{\textbf{Arrangement of primitive objects}.}
As shown in Figure~\ref{fig:arrangement}, we place a collection of $M$ primitive objects from set $G$ into a $3D$ volume $F \in \mathbb{R}^{W \times H \times D}$ in descending order of their volume.
Initially, the elements of the primitive objects set $G$ are sorted in descending order according to the volumes of the primitive objects and are re-indexed as $l'$ to reflect the sorted order.
\ryu{We set a condition regarding overlaps for the arrangement of primitive objects. When placing the object $I_j$ in 3D volume $F$, we define the area already occupied by the objects $\{I_{l'}\}_{l'=1}^{j-1}$ as $A_j$, and the area that $I_j$ occupies as $B_j$. The condition for overlap stipulates that the volume overlap ratio, represented as $\frac{O(A_j \cap B_j)}{O(B_j)}$, should be less than a threshold $r$. If this condition is met, we proceed with the placement. In this context, $r$ denotes the maximum overlap ratio of the shapes, and $O$ is a function representing the volume of the occupied region.}
\ryu{The placement procedure for each primitive object comprises two main steps: (1) position selection and  (2) arrangement. 
(1) position selection: randomly select a position of the center of the primitive object in the $3D$ volume $F$.
(2) arrangement: placing the primitive object based on the overlap condition shown in Figure~\ref{fig:arrangement}. If the overlap condition is met at the position from (1), we place the primitive object. 
If not, we revert to (1).
This (1) and (2) process is repeated up to a maximum of $Max$\_$iter$ (=100). If the (1) and (2) process fails after 100 iterations, the primitive object is rejected.}
\ryu{We repeat the placement process for each object $I_{l'}$ a total of $M$ times, proceeding sequentially from $l'=1$ to $l'=M$. }
The outcome of arranging the contours of the primitive objects is denoted as the assembled object $S_i$. The outcome of placing the primitive objects filled in the interior is denoted as $m_i$. 
\ryu{In this study, we set the intensity values of the contours for the primitive objects within the assembled object to a fixed value, 
$Intensity$.}
We iteratively generate pairs $(S_i, m_i)$ for $N$ times, thereby automatically creating the pre-training dataset $\mathcal{D}$ for PrimGeoSeg.
Combining various random parameters increases the diversity of geometric shapes within each class. We call this intra-class diversity of shapes instance augmentation.
\ryu{Please refer to the Supplementary Materials for the parameters and values required to generate the pre-training data for PrimGeoSeg.}

\subsection{Hypothesis and Motivation of PrimGeoSeg}
\label{sec:intuition}
\textbf{Reasons for independent rules in the $xy$-plane and $z$-axis:}
3D medical images are created by constructing of $xy$ slices and reconstructing them along the $z$-axis.  We thus generate primitive objects by stacking slice images on the $xy$-plane according to the $z$-axis rule.


\noindent{\textbf{Why is each $xy$-plane and $z$-axis rule defined as described above?:}}
3D general object recognition recognizes diverse and complex 3D objects in the real world. On the other hand, 3D medical image segmentation recognizes only a limited number of 3D objects within the human body's internal anatomy. Therefore, in PrimGeoSeg, we considered that a certain number of primitive objects should be sufficient for 3D image segmentation.

\noindent{\textbf{Why introduce the overlap when arranging primitive objects?} }
The internal structure of a human being is such that blood vessels can penetrate the interior of organs. In this case, 3D medical images are represented as if the blood \cRyo{vessels overlap} a part of the organ region. Therefore, we introduce the overlap when arranging multiple primitive objects to create an assembled object in which \cRyo{the part region that overlaps} more closely resembles the internal structure of the human body.

\section{Experiments}

\subsection{Experimental Settings}
\label{imple}
\noindent\textbf{Datasets}.
\cRyo{In this experiment, we evaluate the effectiveness of PrimGeoSeg using several datasets: BTCV~\cite{BTCV},  MSD~\cite{MSD}, and the 2021 edition of the Multi-modal Brain Tumor Segmentation Challenge (BraTS). BTCV has 30 samples for organ segmentation, and we split the BTCV training data in an 8:2 ratio for offline evaluation as in~\cite{chen2023masked}. MSD has the lung (Task06) with 63 samples, and the spleen (Task09) has 41 samples for organ segmentation. In the MSD, we focused on the lung (Task06) and the spleen (Task09) due to computational resource constraints, and we split the training data in an 8:2 ratio for offline evaluation as in ~\cite{ye2022desd}.
BraTS has 1,251 samples for brain tumor segmentation, we performed segmentation of three types of tumors: whole tumor (WT), tumor core (TC), and enhancing tumor (ET), splitting the training data in an 8:2 ratio for offline evaluation\cRyo{, as follows}~\cite{hatamizadeh2022swin}.
}

\noindent\textbf{Architectures}.
We utilized the prominent transformer-based models, UNETR~\cite{hatamizadeh2022unetr} and SwinUNETR~\cite{hatamizadeh2022swin}, as architectures for 3D medical image segmentation. Both UNETR and SwinUNETR have demonstrated their state-of-the-art performance on test leaderboards for \cRyo{the} BTCV and MSD, surpassing the capabilities of conventional CNN-based models. 

\noindent\textbf{Implementation details}.
\ryu{PrimGeoSeg executes the segmentation task with assembled objects $S_i\in\mathbb{R}^{96 \times 96 \times 96}$ as input data and the mask $m_i\in\mathbb{R}^{96 \times 96 \times 96}$ as ground truth in pre-training.}
\ryu{During pre-training of PrimGeoSeg,} we \cRyo{used} $96\times96\times96$ patches, a batch size of 8, a learning rate of 0.0001, and a weight decay of 0.00001, optimizing the dice loss. We employ AdamW~\cite{loshchilov2017decoupled} with a warmup cosine scheduler for training. \ryu{For the number of iterations, when the pre-training data is $\{5K,50K\}$, the iterations are set to $\{100K,375K\}$.} Concerning the pre-training data for PrimGeoSeg, constructing a dataset of 5,000 objects requires less than two hours on a 400GiB CPU memory system, and the storage used is under 3GB. The pre-training process on NVIDIA A100 GPUs takes up to five GPU days for 100,000 iterations. 
For fine-tuning on BTCV, MSD, and BraTS, we follow the conditions of the hyperparameters on each fine-tuning dataset. For specific hyperparameter settings, we refer readers to the respective conventional research. Specifically, for SwinUNETR in BTCV and BraTS, refer to ~\cite{tang2022self}; and for UNETR in BTCV and MSD, consult ~\cite{hatamizadeh2022unetr}. In addition, please see ~\cite{hatamizadeh2022unetr} in terms of MSD. 
All experiments for downstream tasks are conducted using a dice similarity coefficient (Dice) as an evaluation metric. For more detailed settings, please refer to the supplementary materials.

\subsection{Fundamental Experiments (see Table~\ref{tab:exp_LEGOsdb})}
\label{expl}

Fundamental experiments aim to clarify the effectiveness of PrimGeoSeg pre-trained model. 
We focus on five aspects: (a) effects of volumetric shapes, (b) effects of the number of classes, (c) effects of instance augmentation (IA), (d) effects of overlap, and (e) effects of dataset size. 
We pre-trained UNETR using 2,500 data of PrimGeoSeg for all experiments.

\noindent{\textbf{Effects of volumetric shapes:}}
This fundamental experiment (a) aims to compare the effectiveness of pre-training between planar shapes and volumetric shapes. 
\begin{wrapfigure}{r}{0.3\textwidth}
    \vspace{-10pt}
    \includegraphics[width=1.0\linewidth]{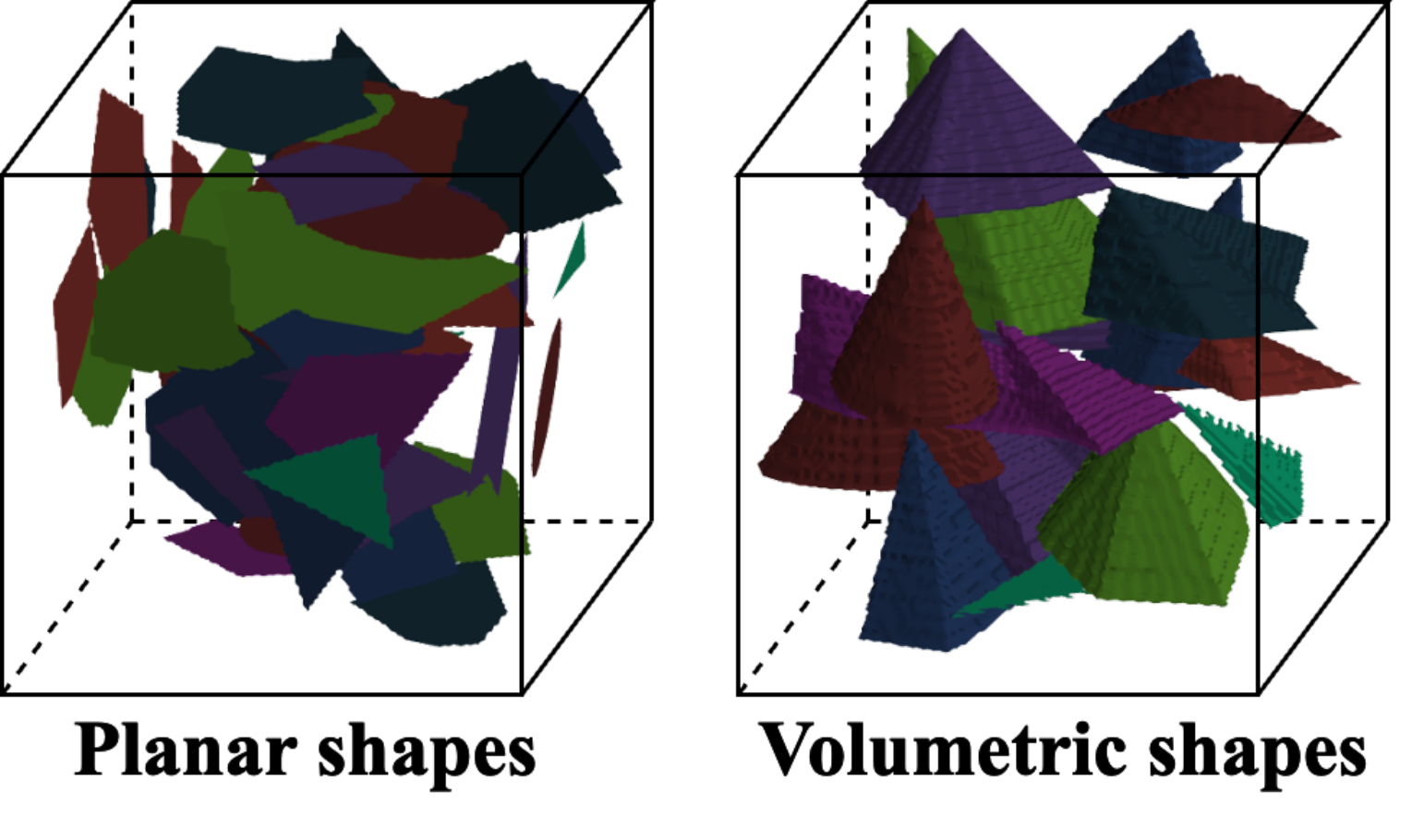}
    \vspace{-23pt}
    \caption{\ryu{Expetiment (a).}}
    \label{fig:fundamental1}
    \vspace{-10pt}
\end{wrapfigure}
We compare the effects of pre-training when arranging planar shapes and volumetric shapes in 3D space \ryu{as shown in Figure~\ref{fig:fundamental1}. }
Table~\ref{tab:exp1} shows that volumetric shapes improve the Dice metrics by 7.78 points compared to planar shapes. This result demonstrates that incorporating volumetric information leads to an improved performance of the pre-training for 3D medical image segmentation. 

\noindent{\textbf{Effects of the number of classes:}} 
This fundamental experiment (b) aims to investigate the validity of each $xy$-plane and $z$-axis rule of our generation method. Table~\ref{tab:exp2} shows that both $xy$ and $z$ rules contribute to the effective pre-training. Moreover, the pre-training effect improves as the number of classes increases. For example, a maximum performance difference of +2.4 points was observed at 1 class and 32 classes. 
This result shows that the pre-training effect is enhanced even for primitive shapes when increasing diversity in the shape of the class. 
We clarify that the diversity of shapes in the xy-plane and z-axis directions in the 3D structure are both important factors in improving the pre-training effect.

\begin{table*}
    \caption{\textbf{Fundamental experiments for BTCV.} Each table show: (a) Effects of volumetric shapes, (b) Effects of the number of classes, (c) Effects of instance augmentation (IA), (d) Effects of overlap, and (e) Effects of the number of pre-training data.}
    \label{tab:exp_LEGOsdb}
    \scriptsize
    \begin{subtable}[t]{0.19\textwidth}
        \centering
        \caption{Shapes}
        \vspace{-5pt}
        \label{tab:exp1}
        \begin{tabular}{lc} \toprule[0.8pt]
                        & Dice \\\midrule[0.6pt]
            Planar      & 69.11 \\
            Volumetric  & \textbf{76.89} \\\bottomrule[0.8pt]
        \end{tabular}
    \end{subtable}
    \hfill
    \begin{subtable}[t]{0.19\textwidth}
        \centering
        \caption{Classes}
           \vspace{-5pt}
        \label{tab:exp2}
        \begin{tabular}{lc} \toprule[0.8pt]
                      & Dice \\\midrule[0.6pt]
            $xy$:1, $z$:1 & 75.12 \\
            $xy$:1, $z$:4 & 76.00 \\
            $xy$:8, $z$:1 & 76.89 \\
            $xy$:8, $z$:4 & \textbf{77.52} \\\bottomrule[0.8pt]
        \end{tabular}
    \end{subtable}
    \hfill
    \begin{subtable}[t]{0.19\textwidth}
        \centering
        \caption{IA}
           \vspace{-5pt}
        \label{tab:exp3}
        \begin{tabular}{lc} \toprule[0.8pt]
               & Dice \\\midrule[0.6pt]
            w/o IA & 74.21 \\
            w IA & \textbf{77.52} \\\bottomrule[0.8pt]
        \end{tabular}
    \end{subtable}
    \hfill
    \begin{subtable}[t]{0.19\textwidth}
        \centering
        \caption{Overlap}
           \vspace{-5pt}
        \label{tab:exp4}
        \begin{tabular}{lc} \toprule[0.8pt]
               & Dice \\\midrule[0.6pt]
            w/o overlap & 77.52 \\
            w overlap & \textbf{78.14} \\\bottomrule[0.8pt]
        \end{tabular}
    \end{subtable}
    \hfill
    \begin{subtable}[t]{0.19\textwidth}
        \centering
        \caption{Dataset size}
           \vspace{-5pt}
        \label{tab:exp5}
        \begin{tabular}{lc} \toprule[0.8pt]
             & Dice \\\midrule[0.6pt]
            0.8K     & 77.39 \\
            2.5K     & 78.14 \\
            50K    & \textbf{80.86} \\\bottomrule[0.8pt]
        \end{tabular}
    \end{subtable}
         \vspace{-10pt}
\end{table*}

\noindent{\textbf{Effects of instance augmentation (IA):}}
This fundamental experiment (c) aims to examine the pre-training effect of our proposed IA method, as it considers individual primitive object variations similar to 3D medical images. As demonstrated in Table~\ref{tab:exp3},  IA improves +3.31 points compared to the performance when not using IA pre-training performance. 
Also, while IA improves accuracy by 3.31 points, class diversity enhances 2.4 points (Table~\ref{tab:exp2}). 
This result suggests that intra-class shape diversity holds equal or greater importance than inter-class shape diversity in pre-training for 3D medical image segmentation.

\noindent{\textbf{Effects of overlap:}}
This fundamental experiment (d) aims to investigate the pre-training effect of overlapping among 3D primitive objects. Because we hypothesize that overlaps between primitive objects could assist pre-training performance by considering\cRyo{, for example,} fuzzy boundaries and overlapping regions within the human body. Table~\ref{tab:exp4} shows that overlapping 3D volumetric shapes led to \cRyo{a} higher accuracy of +0.62 points. 
\ryu{This result suggests that the overlap between primitive objects contributes toward improving the pre-training effect of 3D medical image segmentation. }

\noindent{\textbf{Effects of dataset size:}}
\label{pretrain_scaling}
One of the key advantages of PrimGeoSeg is its ability to generate primitive geometric objects automatically, which enables easily scaling of the pre-training dataset.  As demonstrated in Table~\ref{tab:exp5}, there is a positive correlation between the amount of pre-training data and the effectiveness of pre-training, where more data leads to better pre-training outcomes.
 Note that this experiment is limited to a certain amount of data size due to computational resource constraints.

The above experimental results revealed that volumetric shape extensibility, the number of classes, IA, overlap between primitive objects, and data scalability in PrimGeoSeg contribute to pre-training effects. The above fundamental experiments offered valuable insights into the key elements essential for pre-training in 3D medical image segmentation.

\begin{figure*}[t]
    \centering
    \includegraphics[width=1\linewidth]{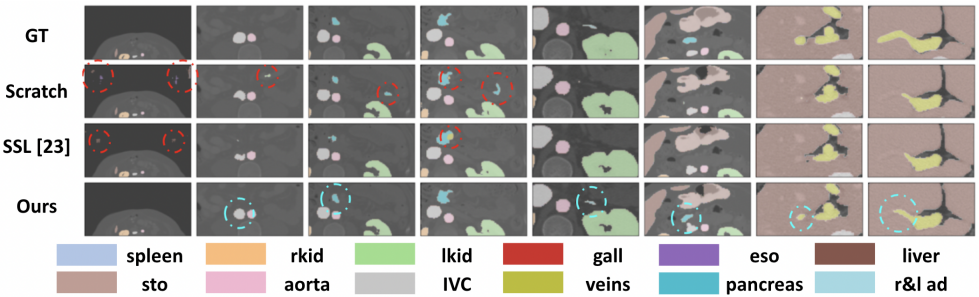}
    \caption{
    \cRyo{
   \textbf{Qualitative results on BTCV.} Red dashes indicate misidentified areas and blue dashes indicate more accurately identified areas.
    }
    } 
  \vspace{-14pt}  
    \label{fig:mask_visu}
\end{figure*}

\subsection{Organ and Tumor Segmentation (see Table~\ref{tab:result_segmentation} and Figure~\ref{fig:mask_visu})}
\label{main}
\ryu{In this section, we verify the effectiveness of PrimGeoSeg on organ segmentation (BTCV and MSD Task09) and tumor segmentation (MSD Task06 and BraTS).}
We employed ~\cite{chen2023masked,tang2022self} of the state-of-the-art SSL only because of limited computational resources and non-integrality of various conditions such as test data, architecture and input size. 

\begin{table*}[t]
    \caption{
    \textbf{3D medical image segmentation benchmark datasets.} The results for PrimGeoSeg on BTCV, MSD, and BraTS in comparison to the previous SSL. The best value for each fine-tuning dataset is in bold.
    }
    \label{tab:result_segmentation}
    \begin{subtable}[t]{0.99\textwidth}
        \centering
        \caption{Comparison of performance in BTCV.}
        \vspace{-4pt}
        \label{tab:result_btcv}
        \setlength{\tabcolsep}{3pt}
        \scalebox{0.74}{
            \begin{tabular}{l|c|c|c|ccccccccccccc}  \toprule[0.8pt]
                Pre-training & PT Num & Type & Avg.  & Spl & RKid & LKid & Gall & Eso & Liv & Sto & Aor & IVC & Veins & Pan & rad & lad \\ \midrule[0.5pt]
                \multicolumn{17}{l}{~\textit{UNETR}} \\
                Scratch  &  0  & --   & 73.0 & 90.2 & 91.1 & 90.7 & 47.0 & 63.8 & 95.3 & 76.5 & 85.1 & 82.1 & 67.9 & 72.3 & 46.1 & 40.8 \\ 
                Chen \textit{et al.}~\cite{chen2023masked} & 0.8K & SSL & 75.8 & 95.2 & \textbf{95.5} & 93.8 & 51.9 & 52.3 & \textbf{98.8} & 80.0 & 87.8 & 82.7 & 66.1 & 68.9 & 60.8 & 51.3 \\ 
                \rowcolor[gray]{0.9} PrimGeoSeg  & 0.8K  & FDSL & 77.4 & 88.9 & 94.0 & 93.8 & 59.8 & 65.7 & 95.4 & 79.3 & 88.3 & 82.6 & 69.9 & 76.8 & 58.5 & 53.3 \\ 
                \rowcolor[gray]{0.9} PrimGeoSeg  & 50K  & FDSL & \textbf{80.9} & \textbf{95.7} & 94.2 & \textbf{94.1} & \textbf{61.9} & \textbf{69.6} & 96.7 & \textbf{85.5} & \textbf{89.5} & \textbf{84.4} & \textbf{74.7} & \textbf{81.9} & \textbf{64.3} & \textbf{58.7} \\ \midrule[0.5pt]
                \multicolumn{17}{l}{~\textit{SwinUNETR}} \\
             Scratch   &  0  & -- & 78.3 & 92.3 & 93.2 & 93.8 & 55.9 & 61.3 & 94.0 & 77.0 & 87.5 & 80.4 & 74.2 & 76.1 & 68.8 & 63.6 \\ 
                Tang \textit{et al.}~\cite{tang2022self} &  5K & SSL & 81.6 & 95.3 & 93.2 & 93.0 & \textbf{63.6} & 74.0 & 96.2 & 79.3 & \textbf{90.0} & 83.3 & \textbf{76.1} & 82.3 & \textbf{69.0} & \textbf{65.1} \\ 
                \rowcolor[gray]{0.9} PrimGeoSeg  &  5K  & FDSL & \textbf{82.0} & \textbf{95.7} & \textbf{94.4} & \textbf{94.4} & 61.0 & \textbf{75.5} & \textbf{96.7} & \textbf{83.3} & 89.1 & \textbf{85.6} & 75.2 & \textbf{84.3} & 67.9 & 62.4 \\\bottomrule[0.8pt]
            \end{tabular}
        }
    \end{subtable}
    \vspace{5pt}
    \\
    \begin{subtable}[t]{0.4\textwidth}
        \centering
        \caption{Comparison of performance in MSD.}
        \label{tab:result_msd}
        \vspace{-4pt}
        \setlength{\tabcolsep}{3pt}
        \scalebox{0.7}{
            \begin{tabular}{l|c|cc|ccc} \toprule[0.8pt]
                                          &         & \multicolumn{2}{c|}{UNETR}  & \multicolumn{2}{c}{SwinUNETR}   \\
                Pre-training & Type & Lung & Spleen & Lung & Spleen         \\\midrule[0.5pt]
                Scratch    &  --   &  52.5   & 95.0             & 63.5   & 96.3                 \\ 
                Tang \textit{et al.}~\cite{tang2022self}  &  SSL    &  --      & --                & 65.2   & 96.5                 \\ 
                \rowcolor[gray]{0.9} PrimGeoSeg  & FDSL &  \textbf{62.2} & \textbf{96.3} & \textbf{67.9} & \textbf{96.6} \\\bottomrule[0.8pt]
            \end{tabular}
        }
    \end{subtable}
    \hfill
    \begin{subtable}[t]{0.57\textwidth}
        \centering
        \caption{Comparison of performance in BraTS.}
        \label{tab:result_brats}
        \vspace{-4pt}
        \setlength{\tabcolsep}{3pt}
        \scalebox{0.7}{
            \begin{tabular}{l|c|cccc|cccc} \toprule[0.8pt]
                                          &         & \multicolumn{4}{c|}{UNETR} & \multicolumn{4}{c}{SwinUNETR}    \\
                Pre-training & Type & Avg.  & ET    & WT    & TC    & Avg.  & ET    & WT    & TC    \\\midrule[0.5pt]
                Scratch      &    --    & 88.1 & 84.8 & 91.3 & 88.1 & 90.0 & 86.8 & \textbf{92.9} & 90.3 \\ 
                \rowcolor[gray]{0.9} PrimGeoSeg    & FDSL &  \textbf{88.7} & \textbf{85.6} & \textbf{91.8} & \textbf{88.9} & \textbf{90.3} & \textbf{87.0} & \textbf{92.9} & \textbf{91.0} \\\bottomrule[0.8pt]
            \end{tabular}
        }
    \end{subtable}
     \vspace{-10pt}
\end{table*}

\noindent\textbf{BTCV.} 
In Table~\ref{tab:result_btcv}, we compare the fine-tuning results of our proposed method, learning from scratch, and the recent state-of-the-art SSL~\cite{chen2023masked,tang2022self}, respectively. 
PrimGeoSeg showed an overall higher recognition performance than Scratch for each class. Even when utilizing an equivalent volume of pre-training data as with the SSL, we observed  performance improvements: UNETR increased by 1.6 points and SwinUNETR by 0.4 points in average Dice score. Moreover, as detailed in Section~\ref{pretrain_scaling}, the performance of UNETR continued to improve as we increased the volume of pre-training data. It is worth noting that with only synthetic 3D pre-training data, the performance of our proposed method is superior to that of the baseline. In addition, Figure~\ref{fig:mask_visu} shows several examples of SwinUNETR output results in BTCV. In areas that are over or under-segmented by Scratch and SSL, PrimGeoSeg is able to segment more accurately. Thus, we speculate that the distinct contours of PrimGeoSeg allow for the acquisition of more accurate 3D structual features during pre-training.

\noindent\textbf{MSD}.
Table~\ref{tab:result_msd} shows the accuracy when fine-tuning to MSD (Task06 and Task09). 
For lung segmentation, both UNETR and SwinUNETR, initialized by PrimGeoSeg improved accuracy compared to training from scratch by \ryu{9.7} points and \ryu{4.4} points, respectively. 
In addition, PrimGeoSeg on SwinUNETR showed a \ryu{2.7} points accuracy improvement compared to SSL.
For spleen segmentation, when using PrimGeoSeg on UNETR and SwinUNETR, PrimGeoSeg exhibited accuracy improvements from Scratch of \ryu{1.3} points and \ryu{0.3} points, respectively. PrimGeoSeg on SwinUNETR had a \ryu{0.1} points accuracy improvement compared to SSL. \cRyo{From this result, PrimGeoSeg demonstrates superior pre-training performance without depending on a specific dataset.}

\noindent{\textbf{BraTS.}} We verify the effectiveness of PrimGeoSeg for tumor segmentation. Due to the difficulty in conducting a fair comparison with other pre-training methods, we primarily focused on comparing PrimGeoSeg with Scratch. 
The BraTS results shown in Table~\ref{tab:result_brats}, indicate that PrimGeoSeg improves accuracy by approximately 0.5 points for ET, WT, and TC, respectively. 
Interestingly, although PrimGeoSeg is designed considering the key elements of the human body's internal structure, it has been proven effective for brain tumor segmentation. 
This suggests that PrimGeoSeg is capable of acquiring a 3D structural representation.

\begin{figure*}[t]
    \centering
    \includegraphics[width=0.85\linewidth]{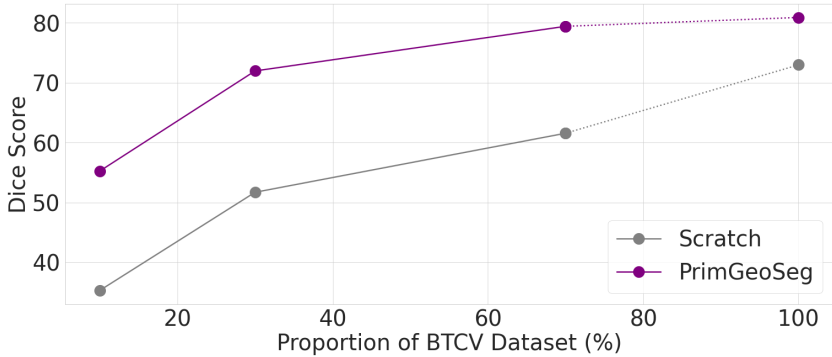}
    \caption{
    \cRyo{
   \textbf{Performance with limited  training data.} 
    } 
  \vspace{-14pt}  
 \label{fig:limited_data}}
\end{figure*}

\subsection{Pre-training Effect on Limited Training Data (see Figure~\ref{fig:limited_data})}
\label{add_exp}


In 3D medical image segmentation, achieving accurate recognition using a small number of 3D medical images is considered ideal. Figure~\ref{fig:limited_data} illustrates the results of limited training data in BTCV. Specifically, this experiment used training data (10\% [2 samples], 30\% [6 samples], 70\% [15 samples]) to compare the performances of Scratch and PrimGeoSeg. 
As shown in Figure~\ref{fig:limited_data},  the accuracy improvements of PrimGeoSeg over Scratch were \{19.9, 20.3, 17.9\}, respectively. Even more surprising, PrimGeoSeg uses only 30\% of the training data, yet it achieves a performance comparable to that of Scratch, which uses 100\% of the training data. This result demonstrates that PrimGeoSeg is beneficial, even with limited training data. Therefore, we consider it a promising pre-training approach to handling limited training data for 3D medical image segmentation.

\section{Conclusion}
\cRyo{
This paper demonstrated the effectiveness of pre-training the proposed PrimGeoSeg, resulting in significant accuracy improvements compared to training from scratch for organ and tumor segmentation. Our proposed method also showed equal or superior performance to self-supervised learning. The findings through experimental results are described below;
}

\noindent\textbf{The effect of the intra-class diversity.} We observed that despite organs in the human body being essentially identical, the size and shape of these organs differ from person to person. In light of this observation, we experimented with a variety of 3D object types and shapes while building PrimGeoSeg. The results demonstrated that increasing the diversity of 3D objects contributes to the enhancement of medical image segmentation performance (see Table~\ref{tab:exp3}).

\noindent\textbf{The effect of spatial overlap of 3D objects.} We also focused on the fact that the anatomical structure of the human body is complex and the boundaries between different tissues and organs are ambiguous, resulting in overlapping regions. Through our exploratory experiments on overlap (see Table~\ref{tab:exp4}), it became clear that incorporating a certain amount of overlap can help improve performance. 

While we empirically confirmed the performance enhancement due to shape pre-training, further analytical justification is required. 
Although the current focus is on addressing data scarcity in segmentation tasks, we plan to investigate the applicability of our method to other domains, such as 3D medical image classification and registration, in future research.


\noindent\textbf{Acknowledgement.} \ryu{
Computational resource of AI Bridging Cloud Infrastructure (ABCI) provided by the National Institute of Advanced Industrial Science and Technology (AIST) was used. 
We want to thank Hideki Tsunashima, Hiroaki Aizawa, Shinagawa Seitaro, and Takuma Yagi for their helpful research discussions.}

\bibliography{egbib}

\includepdf[pages=-]{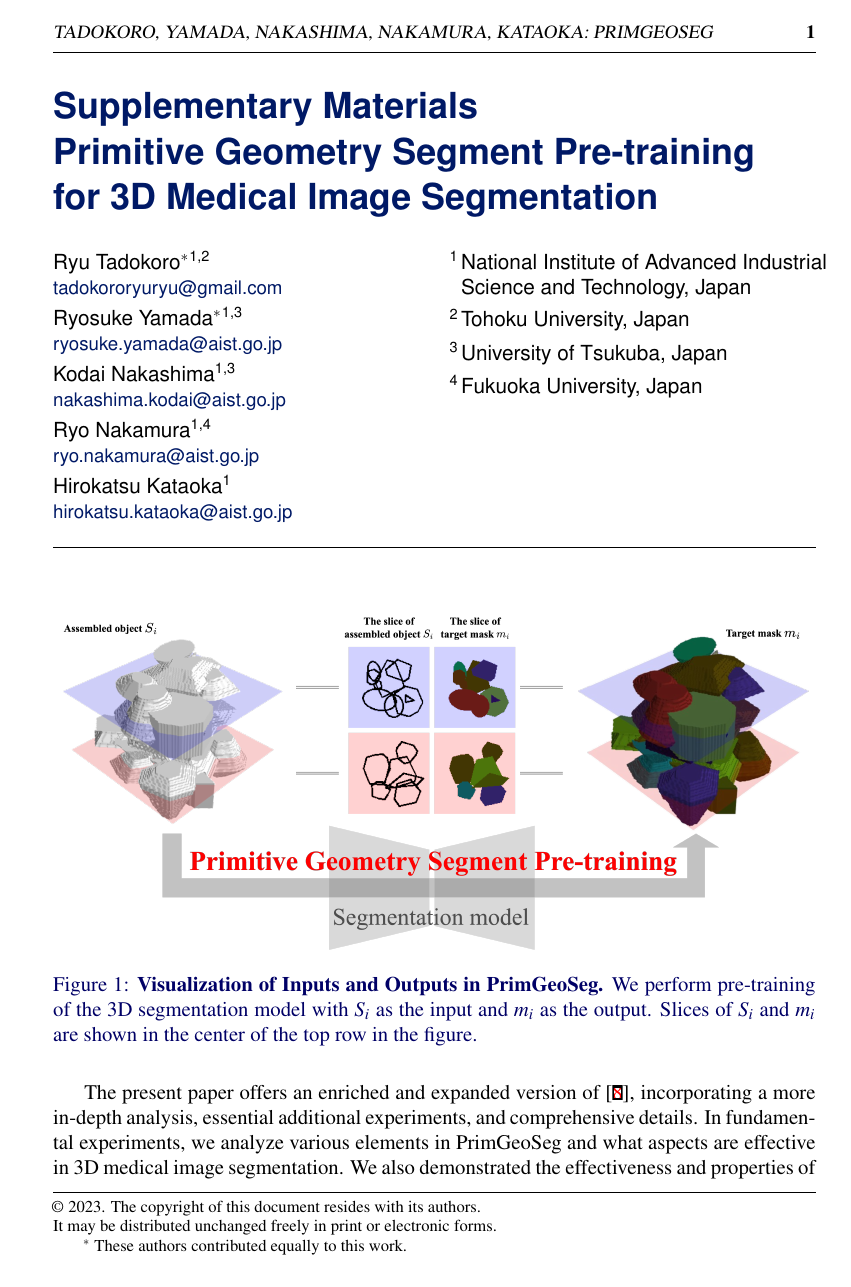}

\end{document}


\maketitle
\blfootnote{$^\ast$ These authors contributed equally to this work.}

\begin{figure*}[h]
  \centering
   \includegraphics[width=1\linewidth]{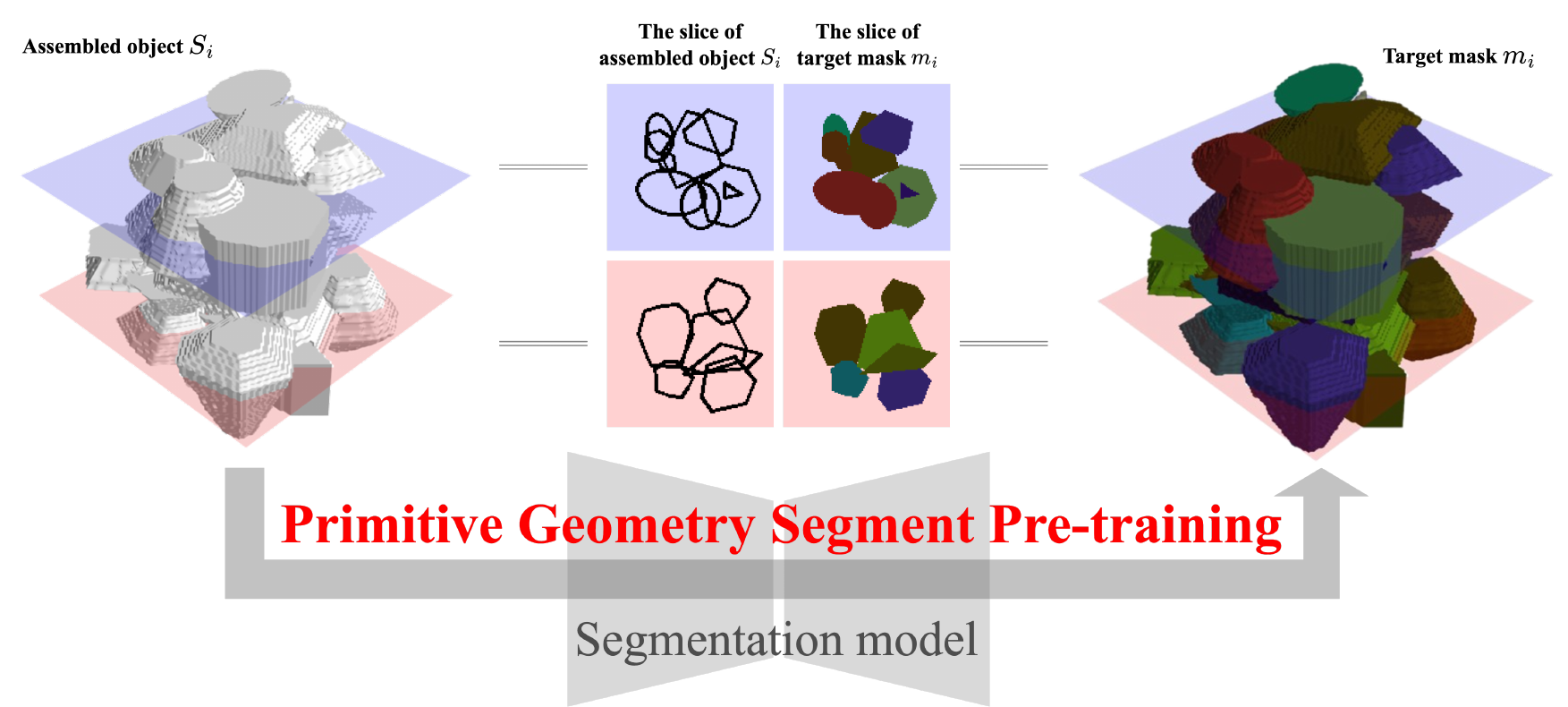}
   \caption{{\textbf{Visualization of Inputs and Outputs in PrimGeoSeg.} We perform pre-training of the 3D segmentation model with $S_i$ as the input and $m_i$ as the output. Slices of $S_i$ and $m_i$ are shown in the center of the top row in the figure.
   }}
   \label{fig:instance_visu}
\end{figure*}
\begin{figure*}[t]
  \centering
   \includegraphics[width=0.9\linewidth]{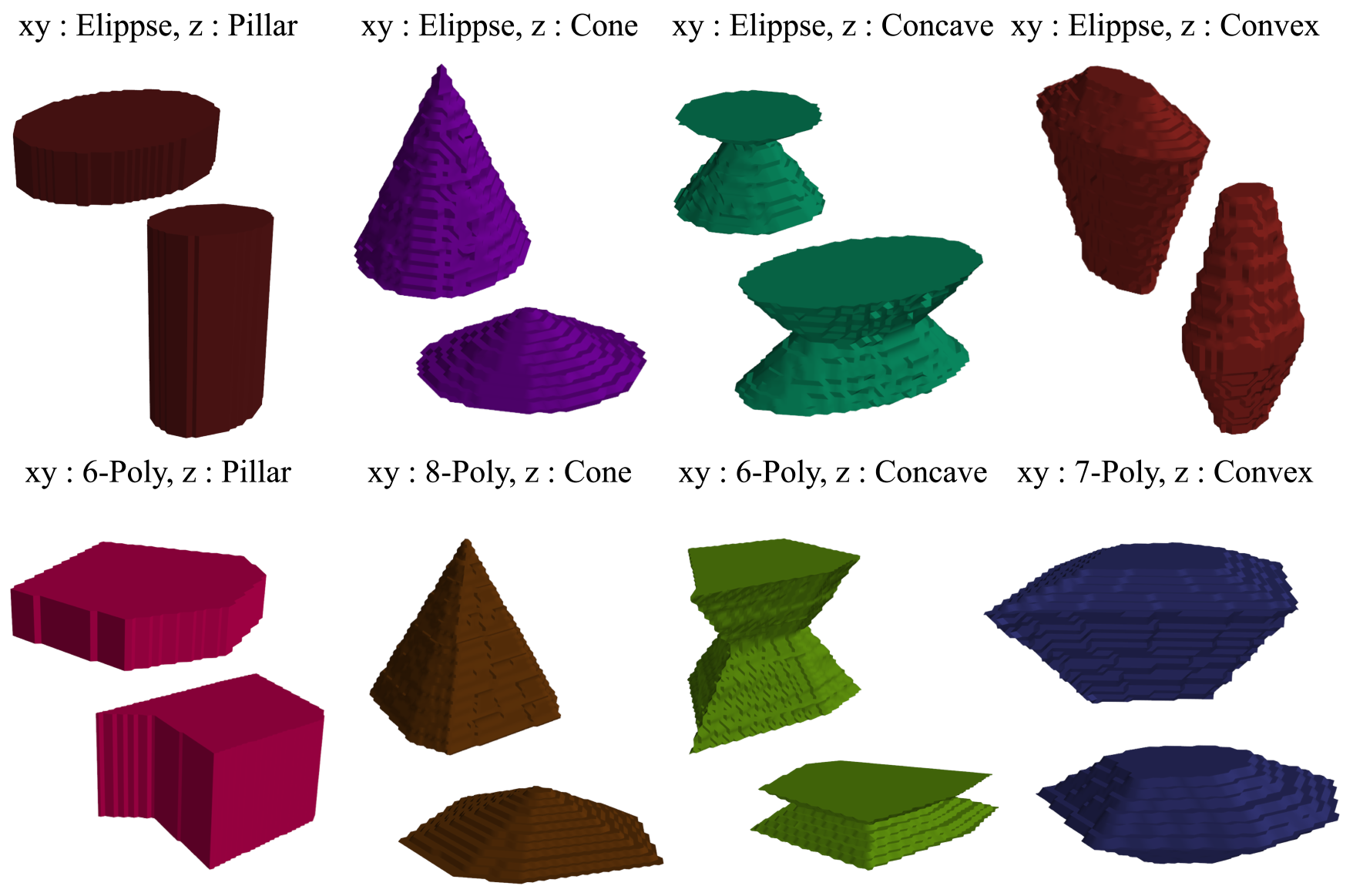}
   \caption{{\textbf{Visualization of Primitive Objects.} Primitive objects are positioned on the assembled object $S_i$. The shape class is uniquely determined by the $xy$-plane and $z$-axis rules. Even within the same class, diversity in shapes is achieved through instance augmentation. 
   }}
   \label{fig:sup_slicevisu}
\end{figure*}

\ryu{The present paper offers an enriched and expanded version of \cite{Tadokoro_2023_CVPR}, incorporating a more in-depth analysis, essential additional experiments, and comprehensive details. In fundamental experiments, we analyze various elements in PrimGeoSeg and what aspects are effective in 3D medical image segmentation. We also demonstrated the effectiveness and properties of shape pre-training in qualitative results, application to a broader range of datasets, and verification of the effect of pre-training on limited data. In this supplementary material, we also delve into further details that were not feasible to encompass within the main manuscript, including an extensive ablation study.} 
In Section~\ref{sec:PrimGeoSeg}, we supplementally provide more detailed information on our proposed PrimGeoSeg. In Section~\ref{sec:benchmark}, we introduce the more detailed experimental settings and the the benchmark dataset on 3D medical image segmentation. \ryu{In Section~\ref{section:add_exp}, we present experimental results which were omitted from the main study due to space constraints.}
\section{Visualization and Details for PrimGeoSeg}
\label{sec:PrimGeoSeg}
\subsection{Input and Output 3D Volumetric Data in PrimGeoSeg}
For PrimGeoSeg, we employ the contour components of primitive objects arranged in 3D space as input for the segmentation task.  The regions filled with primitive objects are target masks. 
Figure~\ref{fig:sup_slicevisu} visualizes the input volumetric image, denoted as $S_i$, and the corresponding target mask, $m_i$.
Presented in the center top of figure~\ref{fig:sup_slicevisu} are the slices of $S_i$ and $m_i$. This visualization confirms that $S_i$ denotes the contour of the primitive objects and $m_i$ functions as a mask, filling the interior of these objects.
Here, we arrange primitive objects in 3D space according to their volume, from largest to smallest. The design of the masks allows subsequent Primitive Objects with smaller volumes to overwrite them.

\subsection{Examples of Primitive objects}
Primitive objects are formulated by integrating rules applicable to the $xy$-plane and the $z$-axis. With eight rules for the $xy$-plane and four for the $z$-axis, we define a total of $32$ shape classes. Examples of these Primitive Objects are illustrated in Figure~\ref{fig:instance_visu}. As can be seen, each unique combination of a $xy$-plane rule and a z-axis rule leads to the specification of a distinct shape class.
Even within the same class, objects may have different shapes. 
This variability results from the random parameters applied during their generation called instance augmentation.
It's important to note that internal human anatomy exhibits individual variation, meaning that even the same organ can have different shapes across individuals.

\begin{figure}[t]
    \centering
    \captionof{table}{\footnotesize Parameters used in the generation of PrimGeoSeg.}
    \includegraphics[width=0.9\linewidth]{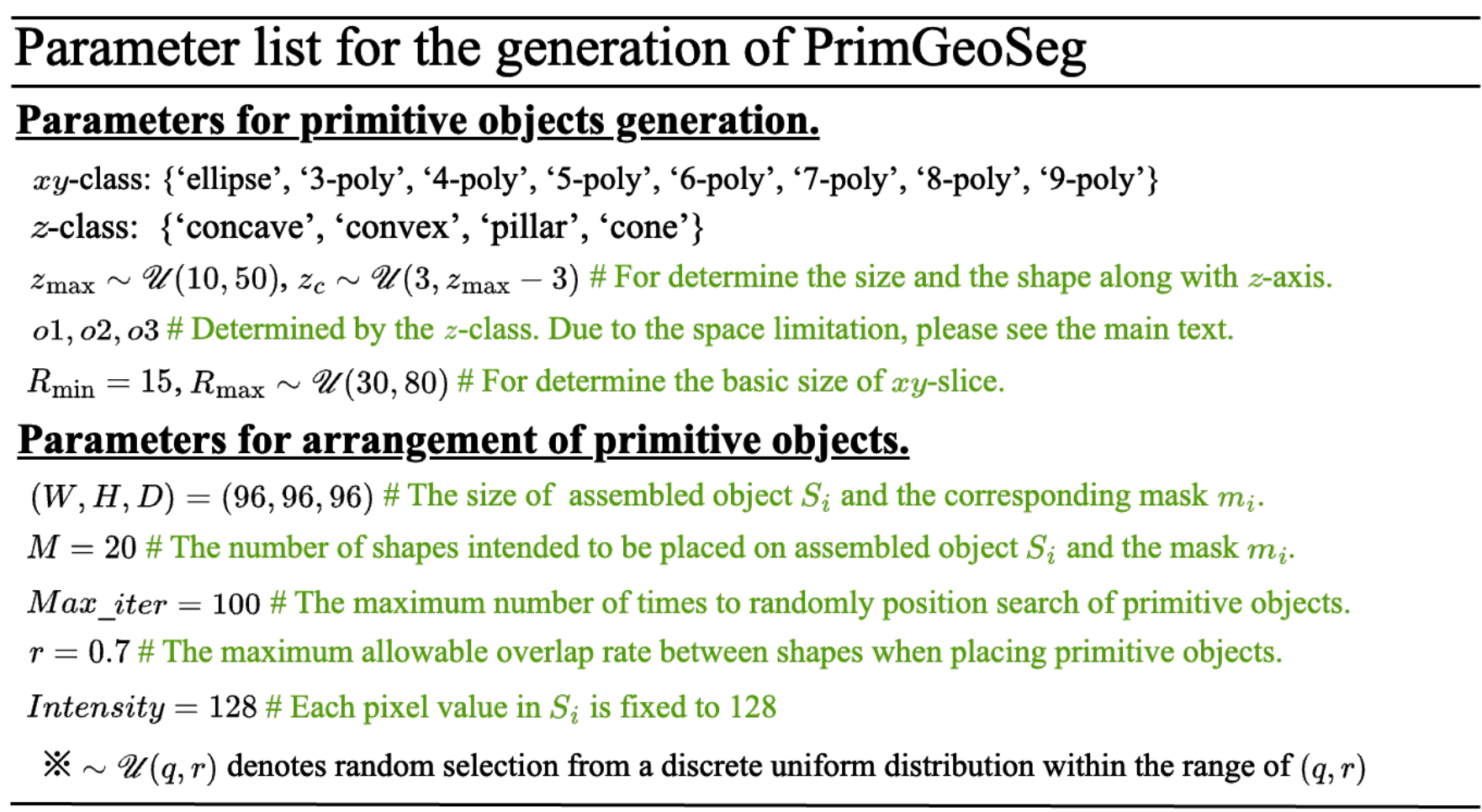}
    \label{tab:parameters}
\end{figure}

\ryu{\subsection{Parameters for the generation process of PrimGeoSeg}
We show the parameters in the whole pipeline of the PrimGeoSeg construction in Table~\ref{tab:parameters}. These parameter values were predetermined to fulfill the motivations outlined in Sec. 3.2 of our paper; no parameter tuning was performed.}

\section{Experimental Details}
\label{sec:benchmark}

\subsection{Datasets for 3D Medical Image Segmentation}
\subsubsection{BTCV}
In our experiments, we use the Multi-Atlas Labeling Beyond the Cranial Vault (BTCV) dataset~\cite{BTCV}, designed for 3D medical image segmentation focusing on human visceral organs. 
The BTCV dataset consists of abdominal CT images from 30 subjects, each meticulously annotated by experts to pinpoint 13 major internal organs. 
Notably, annotations include spleen (Spl), right kidney (RKid), left kidney (LKid), gallbladder (Gall), esophagus (Eso), liver (Liv), stomach (Sto), aorta (Aor), inferior vena cava (IVC), portal vein (Veins), splenic vein (Veins), pancreas (Pan), right adrenal gland (rad), and left adrenal gland (lad).
Every 3D volumetric image, with slice sizes of $512\times 512$ and slice direction, sizes approximately from $100$ to $200$, was resampled into voxels of dimensions $1.5mm\times 1.5mm\times 2.0mm$. We then adjusted each pixel value within the soft tissue window and normalized them to fall within a $[0,1]$ range.
We split the BTCV dataset into \cRyo{train and test set} following an 80:20 ratio, using the test set for offline evaluation. Note that the train and test splits we used are the same as those in the SSL methods we compared with~\cite{chen2023masked, tang2022self}.

\subsubsection{MSD}
The Medical Segmentation Decathlon (MSD) dataset~\cite{MSD} is a comprehensive dataset designed for segmentation tasks across ten different types of tumors and internal organs. 
However, our study focuses on Task06 (Lung) and Task09 (Spleen), given the significant computational \cRyo{cost} demand of analyzing all tasks and the smaller data size of these two tasks for testing data-efficient learning approaches.
Task06 involves performing lung tumor segmentation from 3D volumetric images captured via CT scans, while Task09 requires spleen segmentation from similar 3D CT scan images. We resampled all 3D volumetric images into isotropic voxels of $1.0mm$.
We split the MSD dataset into \cRyo{train and test set} following an 80:20 ratio, using the test set for offline evaluation. 
The results presented in \cite{Tadokoro_2023_CVPR} are given in terms of average dice score for both background and target object. Note that this study reports the Dice score for the target object only. 

\subsubsection{BraTS}
The Multi-modal Brain Tumor Segmentation Challenge (BraTS) dataset~\cite{brats2021} targets identifying Glioblastoma tumor areas captured through MRI \cRyo{images}. 
Four distinct types of multi-modal MRI images—T1-weighted imaging, T1-weighted imaging with contrast enhancement, T2-weighted imaging, and FLAIR imaging—are merged along the channel direction for input to increase the precision in tumor detection. 
The BraTS dataset has annotations for three specific regions: WholeTumor (WT), TumorCore (TC), and EnhancingTumor (ET). 
\cRyo{During training and evaluation, segmentation should be performed for the above three tumor regions.}
We resampled all 3D volumetric images to isotropic voxels of $1.0mm$. Furthermore, we normalize pixel values to achieve a distribution with a mean of 0 and a standard deviation of 1 using non-zero pixel values.
We split the BraTS dataset into \cRyo{train and test set} following an 80:20 ratio, using the test set for offline evaluation.

\begin{table*}[t]
\centering
        \caption{\textbf{Hyperparameters for Each Benchmark Dataset.}}
        \label{tab:hyper_para}
        \setlength{\tabcolsep}{3pt}
        \scalebox{0.90}{
            \begin{tabular}{|l|c|c|c|c|c|c|}
              \hline
              Dataset & \multicolumn{2}{c|}{BTCV} & \multicolumn{2}{c|}{MSD} & \multicolumn{2}{c|}{BraTS} \\
              \hline
              Architecture & UNETR & SwinUNETR & UNETR & SwinUNETR & UNETR & SwinUNETR \\
              \hline
              Optimizer & \multicolumn{6}{c|}{AdamW} \\
              \hline
              Scheduler & \multicolumn{6}{c|}{Warmup cosine scheduler} \\
              \hline
              Input size & \multicolumn{4}{c|}{$96\times 96\times 96$} & \multicolumn{2}{c|}{$128\times 128\times 128$} \\
              \hline
              Batch size & $6$ & $8$ & $6$ & $8$ & $8$ & $8$ \\
              \hline
              Learning rate & \multicolumn{4}{c|}{$0.0001$} & \multicolumn{2}{c|}{$0.0008$} \\
              \hline
              Iteration & $20K$ & $15K$ & $20K$ & $15K$ & \multicolumn{2}{c|}{$125K$} \\
              \hline
            \end{tabular}
        }
\end{table*}

\subsection{Implementation Details}
\ryu{In Section 4.2 of the main study, we conducted five foundational experiments, denoted as (a) through (e). In each experiment, we modified the generation of pre-training data for PrimGeoSeg.
In experiment (a), as shown in Figure 3 of the main study, we defined planar shapes as shapes where each primitive object was positioned with a thickness of 1 in the z-axis direction. For volumetric shapes, in comparison with planar shapes, the class in the z-axis direction was fixed to "Cone", ensuring that there were no overlaps when the shapes were positioned.
In experiment (b), we altered the number of classes for the shapes in PrimGeoSeg. When the class count was set to 1 in the $xy$ plane, we chose "ellipse". In the $z$-axis direction, "cone" was selected for the experiment.
In experiment (c), we compared the pre-training performance based on the presence or absence of instance augmentation for primitive objects. When instance augmentation was disabled, the values for parameters $z_{max}$, $z_c$, $o1$, $o2$, $o3$, and $R_{max}$ presented in Table~\ref{tab:parameters} - were fixed for the experiment.

For downstream tasks of all experiments, we fine-tuned our model using hyper-parameters specifically tailored for BTCV, MSD, and BraTS, as detailed in Table~\ref{tab:hyper_para}. 
Across all experiments, we employ a patch-based approach for both the learning and inference phases. During training, patches of a pre-determined size are randomly cropped from the input images and are then used to train the model. As for the inference stage, we utilize a sliding window technique, with a window overlap of $0.5$, to ensure comprehensive coverage.}

\begin{table}[t]
    \centering
    \caption{Comparison of performance in BTCV.}
    \label{tab:result_btcv}
    \setlength{\tabcolsep}{3pt}
    \scalebox{0.74}{
        \begin{tabular}{l|c|c|c|ccccccccccccc}  \toprule[0.8pt]
            Pre-training & PT Num & Type & Avg.  & Spl & RKid & LKid & Gall & Eso & Liv & Sto & Aor & IVC & Veins & Pan & rad & lad \\ \midrule[0.5pt]
            \multicolumn{17}{l}{~\textit{Swin-based model}} \\
            Scratch  &  0  & --   & 79.5 & \textbf{95.2} & 94.3 & 94.1 & 43.3 & 74.2 & \textbf{96.7} & 78.9 & \textbf{90.2} & 83.5 & 73.1 & 77.7 & 67.5 & \textbf{65.5} \\ 
            Jiang \textit{et al.}~\cite{jiang2022selfsupervised} & 3.6K & SSL & \textbf{81.1} & 93.4 & 94.1 & 94.0 & \textbf{58.5} & 73.7 & 96.3 & \textbf{81.6} & 89.3 & 85.9 & 74.7 & 78.5 & \textbf{68.8} & 65.0 \\ 
            \rowcolor[gray]{0.9} PrimGeoSeg  & 5K  & FDSL & 80.4 & 95.0 & \textbf{94.5} & \textbf{94.4} & 50.1 & \textbf{74.4} & 96.6 & 81.1 & 89.2 & \textbf{86.1} & \textbf{75.9} & \textbf{82.1} & 67.8 & 58.2 \\ \midrule[0.5pt]
            \multicolumn{17}{l}{~\textit{MiT}} \\
         Scratch   &  0  & -- & 78.8 & 93.3 & 93.9 & 93.7 & 62.2 & 70.7 & 96.3 & 77.3 & 86.5 & 80.5 & 72.0 & 73.1 & 64.2 & 61.1 \\ 
            Xie \textit{et al.}~\cite{xie2022unimiss} & 5K+\alpha & SSL & 79.7 & 94.9 & 93.8 & 94.0 & 61.6 & 69.7 & 96.3 & 82.1 & 87.8 & 81.8 & 72.4 & 75.9 & \textbf{66.0} & 60.3 \\ 
            \rowcolor[gray]{0.9} PrimGeoSeg  &  5K  & FDSL & \textbf{82.0} & \textbf{95.4} & \textbf{94.2} & \textbf{94.2} & \textbf{63.6} & \textbf{75.5} & \textbf{96.5} & \textbf{85.7} & \textbf{88.9} & \textbf{85.4} & \textbf{74.7} & \textbf{80.3} & \textbf{66.9} & \textbf{64.9} \\\bottomrule[0.8pt]
        \end{tabular}
        }
\end{table}

\begin{table}[t]
    \centering
    \caption{Comparison of performance in MSD.}
    \label{tab:result_msd}
    \setlength{\tabcolsep}{3pt}
    \scalebox{0.7}{
        \begin{tabular}{l|c|cc|ccc} \toprule[0.8pt]
                                      &         & \multicolumn{2}{c|}{Swin-based model}  & \multicolumn{2}{c}{MiT}   \\
            Pre-training & Type & Lung & Spleen & Lung & Spleen         \\\midrule[0.5pt]
            Scratch    &  --   &  67.4   & 96.5             & 58.6   & 95.8                 \\ 
            Jiang \textit{et al.}~\cite{jiang2022selfsupervised}  &  SSL    &  71.7      & 96.8                & --   & --                 \\ 
            Xie \textit{et al.}~\cite{xie2022unimiss}  &  SSL    &  --      & --                & 70.8   & 95.3                 \\ 
            \rowcolor[gray]{0.9} PrimGeoSeg  & FDSL &  \textbf{73.6} & \textbf{96.8} & \textbf{70.8} & \textbf{95.9} \\\bottomrule[0.8pt]
        \end{tabular}
    }
\end{table}

\section{Additional Experiments}
\label{section:add_exp}
\subsection{Comparison with Other Self-Supervised Learning Methods}
\ryu{The main paper compares SSL techniques using the UNETR and SwinUNETR architectures~\cite{chen2023masked, tang2022self}. 
In this section, we examine the performance of PrimGeoSeg on architectures other than UNETR and SwinUNETR to confirm its generality and effectiveness. We also compared PrimGeoSeg with other self-supervised learning (SSL) methods, namely SMIT~\cite{jiang2022selfsupervised} and UniMiSS~\cite{xie2022unimiss}, using the same architectures for each method. 

SMIT~\cite{jiang2022selfsupervised} employs a Teacher-Student Network with Exponential Moving Averaging for Self-Distillation, optimizing a pseudo-task through Masked Image Modeling. 
For self-supervised pre-training, SMIT uses 3,643 3D CT scans and adopts a 3D segmentation model with the Swin-transformer as its backbone, referred to as the Swin-based model. On the other hand, UniMiSS~\cite{xie2022unimiss} captures multi-modal representations by self-supervised learning on both 3D CT scans and 2D X-ray images simultaneously. UniMiSS utilizes 5,022 3D CT scans and 108,948 2D X-ray images for self-supervised pre-training, incorporating a uniquely designed pyramid U-like medical Transformer (MiT) for its architecture. For simplicity, we trained the Swin-based model and MiT using settings similar to UNETR's. 

Table~\ref{tab:result_btcv} and Table~\ref{tab:result_msd} presents the accuracy comparison when using the Swin-based model and MiT as architectures, with the Dice Score as the evaluation metric. 
Using the Swin-based model, our proposed method, PrimGeoSeg, showed an improvement of 0.9 points in BTCV, 6.2 points in MSD (Lung), and 0.3 points in MSD (Spleen) compared to training from scratch. 
With the MiT, the improvements were 3.2 points in BTCV, 12.2 points in MSD (Lung), and 0.1 points in MSD (Spleen). 

These results suggest that PrimGeoSeg is effective for specific models such as SwinUNETR or UNETR and offers generalizable benefits across other models. 
Compared to the SSL technique SMIT, there was a decrease of 0.7 points in the Average Dice Score for BTCV; however, it performed equivalently or better in half of the classes. 
In MSD (Lung), PrimGeoSeg outperformed SMIT by 1.9 points. Against UniMiSS, PrimGeoSeg showed an improvement of 2.3 points. In the MSD metric, PrimGeoSeg achieved performance on par with UniMiSS. These findings further reinforce the efficacy of our proposed PrimGeoSeg.
}

\begin{table}[t]
    \centering
    \caption{Comparison of Normalized Surface Distance.}
    \label{tab:NSD}
    \setlength{\tabcolsep}{3pt}
    \scalebox{0.7}{
        \begin{tabular}{l|c|ccc|ccc} \toprule[0.8pt]
          &         & \multicolumn{3}{c|}{UNETR}  & \multicolumn{3}{c}{SwinUNETR}   \\
            Pre-training & Type & BTCV & MSD (Lung) & MSD (Spleen) & BTCV & MSD (Lung) & MSD (Spleen)         \\\midrule[0.5pt]
            Scratch    &  --   &  0.715   & 0.511   & 0.899   & 0.766   & 0.673    &   0.953             \\ 
            Tang \textit{et al.}~\cite{tang2022self}  &  SSL    &  --     & --    & --    & 0.829   & 0.686   &   0.960                 \\ 
            \rowcolor[gray]{0.9} PrimGeoSeg  & FDSL &  \textbf{0.822} & \textbf{0.649} & \textbf{0.953} & \textbf{0.839} & \textbf{0.723}   & \textbf{0.967}    \\\bottomrule[0.8pt]
        \end{tabular}
    }
\end{table}

\begin{table}[t]
    \centering
    \caption{The effects of intensity value of PrimGeoSeg.}
    \label{tab:intensity}
    \setlength{\tabcolsep}{3pt}
    \scalebox{0.7}{
        \begin{tabular}{l|c} \toprule[0.8pt]
            Intensity value & BTCV       \\\midrule[0.5pt]
            Pixel value = 128    &  \textbf{81.95}      \\ 
            Pixel value range [78, 178]   & 81.56                  \\\bottomrule[0.8pt]
        \end{tabular}
    }
\end{table}

\subsection{Evaluation using Other Metric}
\ryu{In the main paper, we primarily employed the Dice Score, a prevalent metric in 3D medical image segmentation, for our evaluations. 
In this section, we also evaluate PrimGeoSeg using the Normalized Surface Distance (NSD)~\cite{nikolov2018deep} to provide a more comprehensive assessment. NSD serves as a metric to evaluate the congruence between predicted and ground truth segmentation boundaries, quantifying deviations and providing insights into the precision of boundary delineation. It assigns scores ranging from 0 to 1, where a score closer to 1 indicates a higher congruence between the predicted and actual boundaries. We factored the tolerance threshold of 1mm for each class for NSD. The results of our NSD evaluations, presented in Table~\ref{tab:NSD}, mirrored the trends observed with the Dice Score for both UNETR and SwinUNETR architectures. The enhanced performance evident in both NSD and Dice Score evaluations underscores the effectiveness of our proposed method, PrimGeoSeg.}


\subsection{The Effects of Intensity value of PrimGeoSeg}
\ryu{In the main study, we set the intensity value of the assembled object $S_i$ in PrimGeoSeg to a fixed value of 128. This setting is based on existing findings from shape pre-training research~\cite{KataokaACCV2020}, which indicated that fixing the intensity value led to better pre-training performance. In this section, we examined the pre-training effect when the intensity value was changed from a fixed value to a random value in the range [78,178], using the SwinUNETR architecture and the BTCV dataset. As seen in Table~\ref{tab:intensity}, it is clear that fixing the intensity of the assembled object $S_i$ results in better performance. We believe that fixing the intensity makes the model less focused on texture and more attuned to the shape. Such a focus on shape is crucial for effective shape pre-training.}

\bibliography{egbib}